\documentclass[11pt]{article} 
\usepackage{rldmsubmit,palatino}
\usepackage{graphicx}

\usepackage[utf8]{inputenc}
\usepackage{subfiles}
\usepackage[titletoc]{appendix}
\usepackage{amssymb,amsmath}
\usepackage{amsfonts,mathtools}
\usepackage{url}
\usepackage{xr-hyper}
\usepackage{booktabs} 
\usepackage[breaklinks=true,hidelinks]{hyperref} 
\usepackage{tabulary} 
\usepackage{rotating} 
\usepackage{multirow} 
\usepackage{enumitem}   
\usepackage{mdwlist}	
\usepackage{xspace}
\usepackage{xcolor}
\usepackage{csquotes} 
\usepackage{xmpincl} 
\usepackage{tcolorbox}
\usepackage[ruled,vlined]{algorithm2e}

\newcommand{\notinsubfile}[1]{}
\newcommand{\hangin}{\vspace{5pt}\goodbreak\hangindent=.50cm \noindent }

\newcommand{\vect}[1]{\boldsymbol{#1}}

\title{Swift-Sarsa: Fast and Robust Linear Control}

\author{
Khurram Javed\\
Keen Technologies \thanks{Work done while at the University of Alberta}\\
\texttt{mail@khurramjaved.com} \\
\And
Richard S. Sutton\\
University of Alberta\\
\texttt{sutton@ualberta.ca} \\
}

\begin{document}

\maketitle

\begin{abstract}
Javed, Sharifnassab, and Sutton (2024) introduced a new algorithm for TD learning---SwiftTD---that augments True Online TD($\lambda$) with step-size optimization, a bound on the effective learning rate, and step-size decay. In their experiments SwiftTD outperformed True Online TD($\lambda$) and TD($\lambda$) on a variety of prediction tasks derived from Atari games, and its performance was robust to the choice of hyper-parameters.  In this extended abstract we extend SwiftTD to work for control problems. We combine the key ideas behind SwiftTD with True Online Sarsa($\lambda$) to develop an on-policy reinforcement learning algorithm called \textit{Swift-Sarsa}. 

We propose a simple benchmark for linear on-policy control called the \textit{operant conditioning benchmark}. The key challenge in the operant conditioning benchmark is that a very small subset of input signals are relevant for decision making. The majority of the signals are noise sampled from a non-stationary distribution. To learn effectively, the agent must learn to differentiate between the relevant signals and the noisy signals, and minimize prediction errors by assigning credit to the weight parameters associated with the relevant signals.

Swift-Sarsa, when applied to the operant conditioning benchmark, learned to assign credit to the relevant signals without any prior knowledge of the structure of the problem. It opens the door for solution methods that learn representations by searching over hundreds of millions of features in parallel without performance degradation due to noisy or bad features.  
\end{abstract}

\keywords{
Online learning, Step-size Optimization, SwiftTD
}


\startmain 
\section{Problem Setting}
Our control problem consists of observations and actions. The agent perceives an observation vector $\textbf{x}_t\in \mathbb{R}^n$ at time step $t$. It outputs an action vector $\vect{a}_{t} \in \mathbb{R}^d$. A special component of the observation vector is the reward, $r_t$. The index of the component that is the reward is fixed throughout the lifetime of the agent. Performance on a control problem is measured using the lifetime average reward defined as
\begin{equation}
\begin{aligned}
\text{Lifetime reward}(T) =  \frac{1}{T}\sum_{t=1}^{T} {r}_t. 
\label{controlgoal}
\end{aligned}
\end{equation}
In a control problem, the actions chosen by the agent control what observations the agent perceives in the future, and the agent seeks to maximize its lifetime reward by controlling its future. 
\section{Swift-Sarsa: Fast and Robust Linear Control}
SwiftTD can learn predictions more accurately than prior TD learning algorithms. The ideas that enable it to learn better predictions can be applied to control algorithms as well. The most straightforward way of applying insights from SwiftTD to control problems is to combine its key ideas with True Online Sarsa($\lambda$) (Van Seijen et al., 2016) to develop \textit{Swift-Sarsa}

In our control problem, the output of the agent at every time step is a vector with $d$ components. Swift-Sarsa is limited to problems with a discrete number of actions. If each component of the action vector can only have a finite number of values, then we can represent the problem as having a discrete set of actions.  

Swift-Sarsa uses SwiftTD to learn a value function for each of its $m$ discrete actions. At every time step, it computes the value of each action and stacks them to get an action-value vector. A policy function $\pi : \mathbb{R}^m \rightarrow \{1, \cdots, m\}$ takes as input the action-value vector and returns a discrete action. The value of the action chosen at the current time step is used in the bootstrapped target, and the value of the action chosen at the previous step is used as a prediction when estimating the TD error. The eligibility trace vector of the value function of only the chosen action is incremented.
 
We can make the description of the algorithm more concrete with some notation. Let $\vect{w^i}_t$ be the weight parameter vector for the value function for action $i$ at time step $t$, and let $\vect{\phi_t}$ be the feature vector at time step $t$. The value of the $j$th action is: 
\begin{equation}
 v^j_{t-1, t} = \sum_{i=1}^n w^j_{t-1}[i] \phi_{t}[i].
\end{equation}
The values associated with all actions are stacked to form the action-value vector $\vect{v_{t-1, t}} \in \mathbb{R}^m$ where
\begin{equation}
 {v}_{t-1, t}[j] = v^j_{t-1, t} \text{ for } j \in \{1, \cdots, m\}.
\end{equation}
Let $a_t$ and $a_{t-1}$ be the actions chosen at time step $t$ and $t-1$, respectively. The TD error in Swift-Sarsa is
\begin{equation}
 \delta'_t =  r_t +  \gamma\vect{v}_{t-1, t}[a_{t}] - \vect{v}_{t-2, t-1}[a_{t-1}].
\end{equation}
The eligibility vector for the value function of action $j$ is $\vect{z}^j$. If the action chosen at time step $t$ is $j$ then $\vect{z}^j$ is decayed by $\lambda \gamma$ and incremented using the same update as True Online TD($\lambda$). If the action chosen is different from $j$, then the components of $\vect{z}^j$ are decayed by $\lambda \gamma$ but not incremented. Other than these changes, Swift-Sarsa is the same as SwiftTD. Algorithm \ref{swiftsarsa} is the pseudocode of Swift-Sarsa. 

The policy $\pi$ can be any function. Usually, it is chosen such that actions with high values are more likely to be picked than actions with low values. Two popular choices of policies are the $\epsilon$-greedy policy and the softmax policy. 

The $\epsilon$-greedy policy picks the action with the highest value with probability $1-\epsilon$ and a random action with probability $\epsilon$. The softmax policy turns the values into a discrete probability distribution. The probability of taking the $i$th action is 
\begin{equation}
 \frac{e^{\frac{v_t[i]}{\tau'}}}{\sum_{j=1}^m e^{\frac{v_t[j]}{\tau'}} },
\end{equation}
where $\tau' \in (0, \infty)$ is the \textit{temperature} parameter. Changing the temperature parameter does not change the relative order of likelihood of actions. For a fixed action-value vector, a high value of the temperature parameter makes the policy closer to a uniform policy, and a low value of the temperature parameter makes the policy closer to the greedy policy. In the limit when $\tau' \to \infty$, the probability of every action is the same, and when $\tau' \to 0$, the probability of action with the highest value is one.
\section{The Operant Conditioning Benchmark}
We designed a test bed called \textit{the operant conditioning benchmark} for evaluating the performance of Swift-Sarsa. The benchmark defines a set of control problems that do not need sophisticated strategies for exploration, and a random policy picks the best actions occasionally. The optimal policy for problems from the benchmark can be represented by a linear learner. 

The observation vectors in problems in the benchmark have $n$ binary components, and the action-vectors have $d$ binary components. Both $n$ and $d$ are hyperparameters, and any combination of them for which $n > d$ defines a valid control problem. 

At some special time steps, exactly one of the first $m$ components of the observation vector is one. They are zero on all other time steps. On time steps when the $i$th component of the first $m$ components is one, the agent gets a delayed reward for picking an action-vector whose $i$th component is one and other components are zero. The reward is delayed by $k_1$ steps, where $k_1$ is a variable that is uniformly sampled from $({ISI}_1, {ISI}_2)$ every time the agent picks the rewarding action. On all other time steps, the reward is zero. 

One randomly chosen component from the first $m$ components of the observation vector are one every $k_2$ time steps, where $k_2$ is a variable that is uniformly sampled from $({ITI}_1, {ITI}_2)$

At every step, each of the remaining $n-m$ components of the observation vector is one with probability $\mu_t$. $\mu_1 = 0.05$, and it is recursively updated as
\begin{equation}
 \mu_t = \begin{cases}  \mu_{t-1} + \mathcal{N}(0, 10^{-8}) &\text{if } 0.01 \le  \mu_{t-1} + \mathcal{N}(0, 10^{-8}) \le 0.1\\ 0.01 &\text{if }   \mu_{t-1} + \mathcal{N}(0, 10^{-8}) < 0.01 \\0.1 &\text{if }   \mu_{t-1} + \mathcal{N}(0, 10^{-8}) > 0.1. \end{cases}
\end{equation}
Intuitively $\mu$ is the value of a random walk that starts at 0.05, and it is updated by adding a sample from $\mathcal{N}(0, 10^{-8})$ at every time step. $\mu$ is forced to stay in the range $[0.01, 0.1]$. 

The last $n-m$ components of the observation vector are a source of noise with a time-dependent distribution. Control problems whose observations have many noisy components ($n-m$ is large) are challenging. 

\begin{algorithm}[tbh]
 \DontPrintSemicolon
 \caption{Swift-Sarsa}
 Hyperparameters: $\epsilon = 0.999, \eta = 0.1, \eta^{min} = e^{-15}, \alpha^{init} = 10^{-7}, \gamma, \lambda, \theta$\;
 Initializations: $\vect{w}, \vect{h^{old}}, \vect{h^{temp}}, \vect{z^{\delta}}, \vect{p}, \vect{h}, \vect{z}, \bar{\vect{z}} \leftarrow \vect{0} \in \mathbb{R}^n; (v^{\delta},v^{old}) = (0, 0); \vect{\beta} \leftarrow \text{ln}(\vect{\alpha}^{init}) \in \mathbb{R}^n$\;
 \While{alive}{
 Perceive $\vect{\phi}$ and $r$\;
 \For{$i \in {0, \cdots, m}$}{
    $v[i] \leftarrow \sum_{{\phi}[j] \ne 0} {w^i}[j]{\phi}[j]$\;
 }
    $k \leftarrow \pi(\vect{v})$\;
    $\delta' \leftarrow r + \gamma v[k] - v^{old}$\;
 \For{${z}^j[i] \neq 0 \text{ } \forall i, j$}
 {
        ${\delta^{w^j}}[i] \leftarrow \delta' {z^j}[i] - {z^{\delta^j}}[i]v^{\delta}$\;
        ${w^j}[i] \leftarrow {w^j}[i] + {\delta^{w^j}}[i]$\;
            $\color{red} {\beta^j}[i]  \leftarrow \color{red} {\beta^j}[i] + \frac{\theta}{e^{{\beta^j}[i]}} (\delta' - v^{\delta}) {p^j}[i] $ \;
            $\color{green} {\beta^j}[i] \leftarrow \text{clip}\left({\beta^j}[i], \text{ln}(\eta^{min}), \text{ln}(\eta)\right)$\;

            $\color{red}{h^{old^j}}[i] \leftarrow {h^j}[i]$\;
            $\color{red}{h^j}[i] \leftarrow {h^{temp^j}}[i] + {\delta'} {\bar{z}^j}[i]  - z^{\delta^j}[i]v^{\delta}  $\;
            $\color{red}{h^{temp^j}}[i] \leftarrow {h^j}[i]$\;
        ${z^{\delta^j}}[i] = 0$\;
    $({z^j}[i], \color{red} {p^j}[i], \color{red} {\bar{z}^j}[i]) \color{black} \leftarrow (\color{black} \gamma\lambda {z^j}[i], \color{red} \gamma\lambda {p^j}[i], \gamma\lambda {\bar{z}^j}[i])$\;
    
 }

    $ v^{\delta}  \leftarrow 0$\;
    $\color{blue}\tau \leftarrow  \sum_{{\phi}[i] \ne 0} e^{{\beta^k}[i]}{{\phi}[i]}^2$\;
    $b \leftarrow \sum_{{\phi}[i] \ne 0} {z^k}[i]{\phi}[i]$\;
 \For{${\phi}[i] \neq 0$}{
    $v^{\delta} \leftarrow v^{\delta} + {\delta^{w^k}}[i] {\phi}[i]$\;
    ${z^{\delta^k}}[i] \leftarrow  \color{blue}\text{min}\left(1, \frac{\eta}{\tau}\right)\color{black}e^{{\beta^k}[i]}{\phi}[i]$ \tcp*[]{$\eta$-bound}
    ${z}^k[i] \leftarrow {z^k}[i] + {z^{\delta^k}}[i](1 - b)$\;
    $\color{red} {p}^k[i] \color{black} \leftarrow \color{red} {p}^k[i] + {\phi}[i]{h}^{old^k}[i]$\;
    $ \color{red}{\bar{z}^k}[i] \leftarrow {\bar{z}^k}[i]   + {z^{\delta^k}}[i]\left(1 -  b - {\phi}[i] {\bar{z}^k}[i] \right)$\;
    $\color{red}{h^{temp^k}}[i] \leftarrow {h^k}[i]  - z^{\delta^k}[i]\phi[i]h^k[i] - {h^{old^k}}[i] \phi[i](z^k[i] - z^{\delta^k}[i])  $\;
 \If{$\tau  > \eta$}{
        $\color{brown} {\beta^k}[i] = {\beta^k}[i] + {\phi}[i]^2 \text{ln}(\epsilon)$ \tcp*[]{Step-size decay}
        $\color{brown} ({h^{temp^k}}[i], {h^k}[i], {\bar{z}^k}[i]) = (0, 0, 0)$\;
 }
 }
    
    $v^{old} \leftarrow v[k]$\;
    \label{swiftsarsa}
 }
\end{algorithm}
The inspiration for the operant conditioning benchmark is the \textit{animal learning benchmark} by Rafiee et al. (2023). The animal learning benchmark is inspired by classical conditioning experiments done by behaviorists on animals, and the operant conditioning benchmark is inspired by operant conditioning experiments. The key difference between them is that in operant conditioning experiments, the actions chosen by the animals influence the rates of the rewards. In classical conditioning experiments the animals have no control over the rates of rewards and simply learn to predict the upcoming rewards (\textit{e.g.}, Pavlov's dog).
\section{Experiments: Swift-Sarsa on the Operant Conditioning Benchmark}
We ran experiments with Swift-Sarsa on the operant conditioning benchmark for different values of $n$. We set $d=2$ in all experiments. $({ISI}_1, {ISI}_2)$ was  $(4, 6)$, and $({ITI}_1, {ITI}_2)$ was (50, 70). The lifetime of the agent was 300,000. The policy was softmax with a temperature parameter of $0.1$. The action-vector is mapped to a discrete set. Actions (0, 0), (0, 1), (1, 0), and (1, 1) are mapped to discrete actions 1, 2, 3, and 4, respectively.
\begin{figure}
 \centering
 \includegraphics[width=\linewidth]{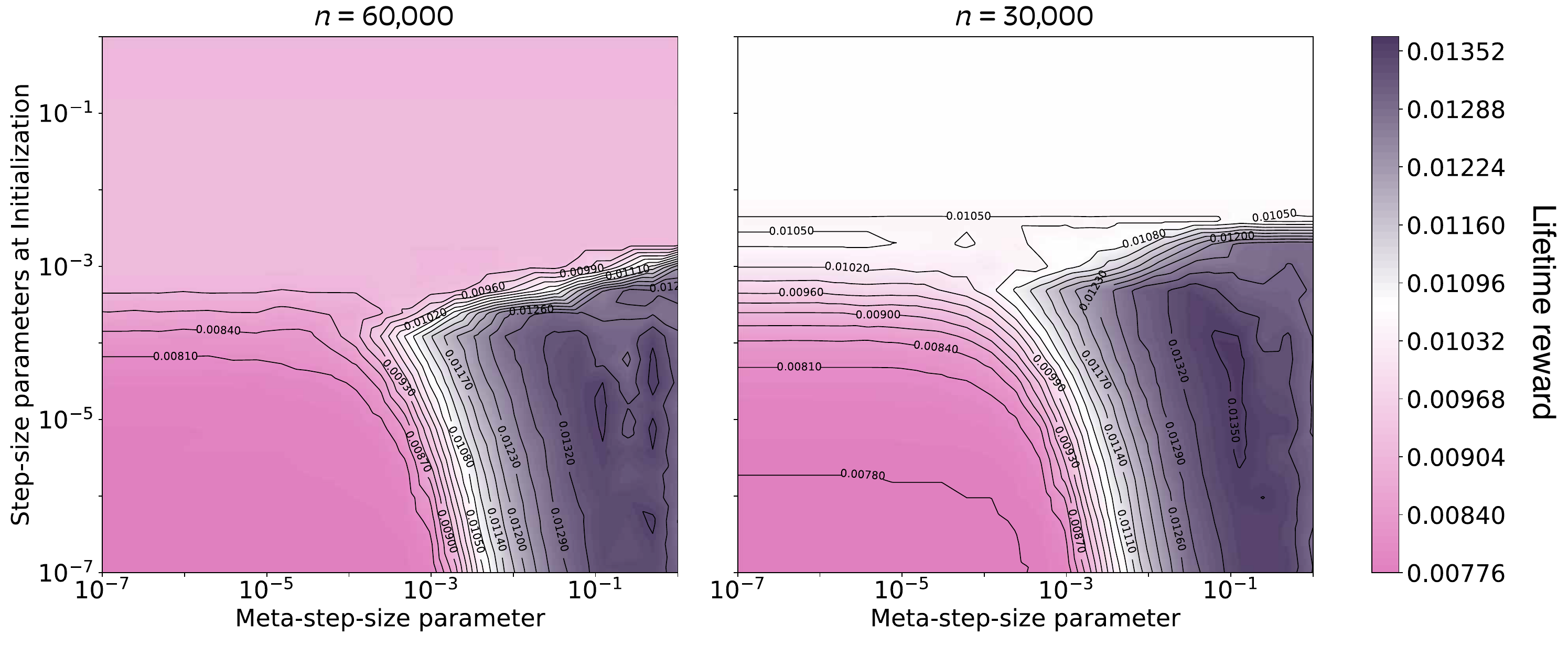}
 \caption{Performance of Swift-Sarsa as a function of the meta-step-size parameter and the initial values of step-size parameters on the operant conditioning benchmark. Experiments in the left figure had $n=60,000$ and the right figure had $n=30,000$. For both set of experiments $\eta$ was $1.0$, $m$ was 2, and $\epsilon$ was $0.9999$. }
    \label{Distractors}
\end{figure}
\begin{figure}
 \centering
 \includegraphics[width=\linewidth]{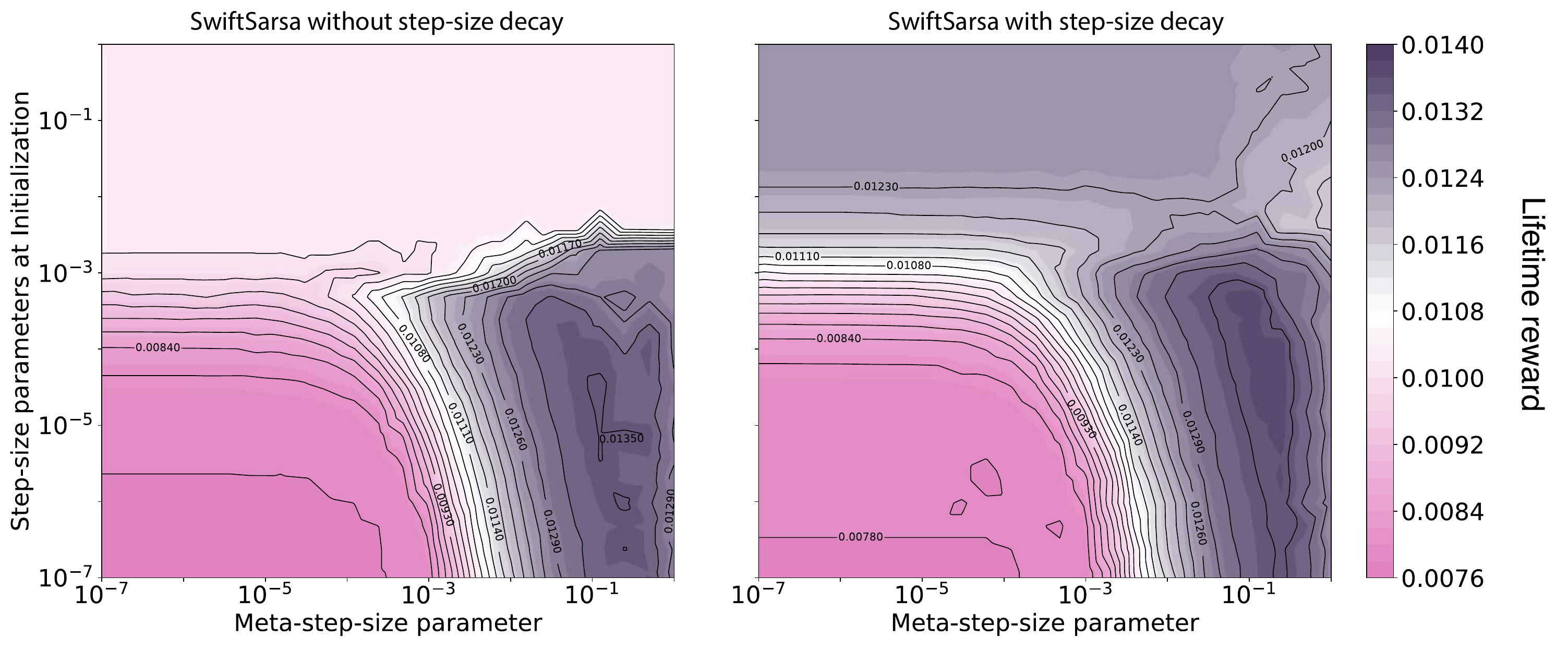}
 \caption{Impact of step-size decay on the performance of Swift-Sarsa as a function of the meta-step-size parameter and the initial values of step-size parameters on the operant conditioning benchmark. Experiments in the left panel did not use step-size decay whereas experiments in the right panel used a step-size decay with decay parameter set to 0.999. Comparing the two results we see that step-size decay improves performance when the initial value of the step-size parameters is too large. For both sets of experiments, $\eta$ was one and $m$ was two.}
    \label{DecayAblation}
\end{figure}

Figure \ref{Distractors} plots the average reward for different values of the meta-step-size parameter and the initial value of the step-size parameters for two different values of $n$. Similar to the performance of SwiftTD, the performance of Swift-Sarsa improved as the meta-step-size parameter increased showing the benefit of step-size optimization. For a wide range of its parameters, Swift-Sarsa achieved a lifetime reward that was close to the optimal lifetime reward of $\approx 0.014$. Increasing the number of distractors made the problem more challenging, and the performance of Swift-Sarsa worsened. 

In a second set of experiments, we compared the impact of step-size decay on the performance of Swift-Sarsa. The results are in Figure~\ref{DecayAblation}. Similar to its impact on SwiftTD, step-size decay improved the performance of Swift-Sarsa when the initial value of the step-size parameters was too large. 

Swift-Sarsa is a simple way of transferring the improvements made by SwiftTD to control problems. A more thorough evaluation of Swift-Sarsa on a wider range of control problems is needed to understand its full potential. It is possible that Swift-Sarsa when combined with more powerful preprocessing, such as \textit{tile coding} (Sutton \& Barto, 2018), can perform similarly to deep RL algorithms on more complex problems, such as Atari games. 
\newpage
\section*{References}
\hangin Javed, K., Sharifnassab, A., \& Sutton, R. S. (2024). Swifttd: A fast and robust algorithm for temporal difference learning. \textit{Reinforcement Learning Journal}.
\hangin Rafiee, B., Abbas, Z., Ghiassian, S., Kumaraswamy, R., Sutton, R. S., Ludvig, E. A., \& White, A. (2023). From eye-blinks to state construction: Diagnostic benchmarks for online representation learning. \textit{Adaptive Behavior}.
\hangin Sutton, R. S., \& Barto, A. (2018). \textit{Reinforcement Learning: An Introduction} (2nd ed.). MIT Press.
\hangin Van Seijen, H., Mahmood, A. R., Pilarski, P. M., Machado, M. C., \& Sutton, R. S. (2016). True online temporal-difference learning. \textit{Journal of Machine Learning Research}.

\end{document}